# Marvelous Agglutinative Language Effect on Cross Lingual Transfer Learning

**Wooyoung Kim**  **Chaerin Jo**  **Minjung Kim**  **Wooju Kim***

Smart System Lab[1]
Department of Industrial Engineering, Yonsei University
{timothy, crjo, mjk0150, wkim}@yonsei.ac.kr

## Abstract

*E-Commerce Services such as Amazon and Alibaba have international consumers. The languages used in those services are very diverse. In order to provide a product search system to international consumers, it is essential to develop artificial intelligence that captures the semantic similarities among different languages. In this study, we propose a model that captures similarities between various languages as one artificial intelligence model. As for multilingual language models, it is important to select languages for training because of the curse of multilinguality. It is known that using languages with similar language structures is effective for cross lingual transfer learning. However, we demonstrate that using agglutinative languages such as Korean is more effective in cross lingual transfer learning. This is a great discovery that will change the training strategy of cross lingual transfer learning.*

## Introduction

E-Commerce Services such as Amazon and Alibaba have international consumers. The languages used in those services are very diverse. In order to provide a product search system to international consumers, it is essential to develop artificial intelligence that captures the semantic similarities among different languages. In this study, we propose a model that captures similarities between various languages as one artificial intelligence model.
We propose a training strategy for cross lingual semantic similarity embeddings using Siamese Network (Reimers and Gurevych., 2019). We finetune the pretrained multilingual transformer, XLM-RoBERTa (Conneau et al., 2020). It focuses on scaling more languages and increasing model capacity, following the common semantic search method to map each sentence to a vector space so that semantically similar sentences are close.

According to Reimers and Gurevych (2019), un- finetuned BERT (Devlin et al., 2018) embeddings rather yield bad sentence embeddings. To mitigate this issue, they utilize the Siamese Network enabling the semantically similar sentences to be found through cosine similarity. Since the structure of XLM-

[1] http://smartweb.yonsei.ac.kr

* corresponding author

RoBERTa (XLM-R) follows the masked language model objective, similar to the one of BERT, it has the same problem. So, we finetune XLM-R through the Siamese Network based on what Reimers and Gurevych (2019) found to find the semantic similarities of cross lingual languages.

Interestingly, our model trained on only Korean dataset outperforms the model trained on English dataset in the evaluation using English test dataset. To explain this phenomenon, we propose a hypothesis that agglutinative languages such as Korean are more effective for cross lingual semantic similarity than isolated languages such as English.

## Related Work

### *XLM-RoBERTa (XLM-R)*

XLM-RoBERTa (XLM-R) is a transformer-based model pretrained on large scale multilingual data (Conneau et al., 2020). XLM-R shows remarkable results on cross lingual transfer learning. However, there is a limitation that the unfinetuned XLM-R cannot fully capture the semantic similarity of sentences. To overcome this issue, we compose XLM-R with the Siamese Network and finetune it with the combination of various languages.

### *Linguistic Typology*

We propose a new perspective of cross lingual learning based on linguistic typology. In linguistic typology, language types are classified into three types - isolated language, inflectional language, and agglutinative language in terms of syntactic stability.[2] The concept of syntactic stability in language indicates that there are some transformations among languages in terms of free word order.

First, isolated languages such as **English** have strict word order, meaning that they have high syntactic stability because word order structure and grammar are directly connected. Second, inflectional languages such as **German** have intermediate characteristics between isolated and agglutinative languages. Lastly, agglutinative languages such as **Korean and Turkish** have low syntactic stability, meaning that they have relatively free word order. According to Choi and Schmitt (2015), the uses of Korean postpositions are different from those in English.

We assume that these grammatical differences among languages have influenced the performance of cross lingual transfer learning. Our experiments indeed demonstrate that the higher free word order, the better performance in cross language learning.

## Experiments

### *Model Architecture*

We train XLM-R by using the Siamese Network (Reimers and Gurevych, 2019). In our experiments, we use a softmax classifier head for Natural Language Inference (NLI) classification task and adopt an average pooling value without the head layer for Semantic Textual Similarity (STS) regression tasks.
In order to finetune XLM-R, we construct the Siamese Network to update the weights of semantically meaningful sentence embeddings. We experiment with the following architecture and objective functions as shown in Figure 1.

[2] There is also incorporating language, but it is too low-resource and difficult to identify distinct characteristics. So, we exclude it from our research.

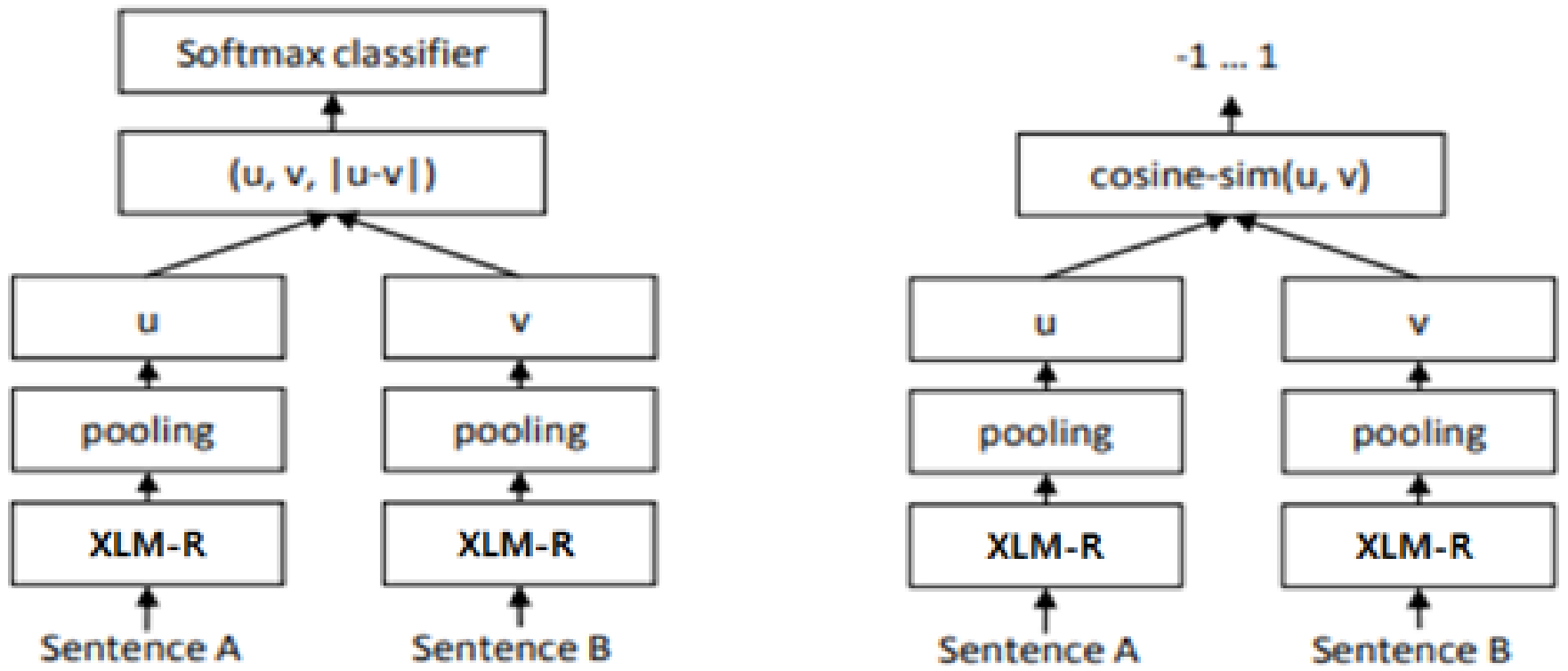

**Figure1. The architecture of XLM-R using Siamese Network. Different sentences are put into Sentence A and B as inputs. The left side is the head version, and the right side is the non-head version.**

### *Loss Function*

Henderson et al. (2017) found that Multiple Negative Ranking Loss is not only much intuitive but significantly better in producing sentence embeddings. So, we use Multiple Negative Ranking Loss. The training data for Multiple Negative Ranking Loss consists of sentence pairs $[(x_1, y_1), \ldots (x_n, y_n)]$ where we assume that $(x_i, y_i)$ are similar sentences and $(x_i, y_j)$are dissimilar sentences for $i \neq j$. The goal is to minimize the distance between $(x_i, y_i)$ while it simultaneously maximizes the distance $(x_i, y_j)$for all $i \neq j$. A set of $k$ possible responses is used to approximate $P(y \mid x)$ : one correct response and $k - 1$ random negatives. For a batch of size $k$, there will be $k$ inputs $x = (x_1, \ldots, x_k)$ and their corresponding responses $y = (y_1, \ldots, y_k)$. Every reply $y_j$ is effectively treated as a negative candidate for $x_i$ if $i \neq j$.

We optimize Multiple Negatives Ranking Loss instead of cross entropy loss. The goal of training is to minimize the approximated mean negative log probability of the data.

### *Training and Testing Data*

We train our model with only NLI dataset and test it with STS dataset, conducting zero shot learning and evaluating how well our model learns semantic textual similarity. In order to comprehend the correlation of cross learning effects, we perform our experiments on isolated language (English), inflectional language (German), and agglutinative language (Korean, Turkish). We use the following datasets in the experiments: Training Dataset consists of MNLI dataset - English, German, Turkish, and Korean. In this section, MNLI, XNLI, and KorNLI have the same meaning, so sentence pairs can be created by crossing over languages.

- MNLI: Multi-Genre Natural Language Inference corpus is crowd-sourced collection of 433k sentence pairs annotated with textual entailment information (Williams et al., 2018).
- XNLI: The pairs are annotated with textual entailment and translated into 14 languages. We select German and Turkish sentences from XLI dataset (Conneau et al., 2018).
- KorNLI: Korean NLI dataset machine- translated from existing English training sets (Ham et al., 2020).

Mono Lingual Pairs are the same as the source dataset without creating additional sentence pairs.

- En: (En, En)
- De: (De, De)
- Tr: (Tr, Tr)
- Ko:(Ko, Ko)

Cross Lingual Pairs are created by crossing over languages without mono lingual languages:

- En-De: (En, De)
- En-Tr: (En, Tr)
- En-Ko: (En, Ko)
- De-Tr: (De, Tr)
- De-Ko: (De, Ko)
- Tr-Ko: (Tr, Ko)

Mono + Cross Lingual Pairs are obtained by combining mono language and cross language pairs:

- En-De: {(En, En), (De, De), (En, De)}
- En-Tr: {(En, En), (Tr, Tr), (En,Tr)}
- En-Ko: {(En, En), (Ko, Ko), (En, Ko)}
- De-Tr: {(De, De), (Tr, Tr), (De, Tr)}
- De-Ko: {(De, De), Ko, Ko), (De, Ko)}
- Tr-Ko: {(Tr, Tr), (Ko, Ko), (Tr, Ko)}

Whole pairs are obtained by merging Mono and Cross pairs in all languages used in our experiments:

- Whole: {(En, En), (De, De), (Tr, Tr), (Ko, Ko), (En, De), (En, Tr), (En, Ko), (De, Tr), (De, Ko), (Ko, Tr)}

STS Benchmark (STSb) Test Dataset is used for evaluating the performance of our model with the combination of various sentence pairs. Semantic Textual Similarity (STS) calculation in cross-lingual setup is used for estimating the degree of the similarity between two sentences. We measure the performance of the model on STSb dataset through cosine similarity and spearman correlation coefficient of human label scores.

- STSb_En: The similarity of two sentences is manually labeled, ranging from 0 to 5 (Cer et al., 2017).
- STSb_De: We obtain German STSb dataset from huggingface. [3]
- STSb_Ko: KorSTS dataset is a translation of STSb_En (Ham et al., 2020).
- STSb_Tr: We translate STS_En into Turkish by using google translation.

Multilingual Tests consist of the following combinations: (En, En), (De, De), (Tr, Tr), (Ko, Ko).
Cross Lingual Tests are conducted by changing the order of two sentences. Therefore, we use the Spearman correlation coefficient for each sentence and calculate the average for both sentences as scores. These tests consist of the following combinations: (En, De), (En, Tr), (En, Ko), (De, Tr), (De, Ko), (Ko, Tr).

### *Experimental Results*

In mono lingual pair experiment, our model trained on only Korean dataset shows better performance on English STS test data than the one trained on English dataset. Even in Turkish dataset, our trained model performed better than the one trained on English. This similar tendency is also found in cross lingual and Mono + Cross lingual pair experiments, which include agglutinative languages in common.

In fact, it is evident in table 1 that the result of Mono + Cross Tr-Ko outperforms the one of whole pairs. This evidence supports our hypothesis that it is important to use the agglutinative languages as training data to improve cross lingual transfer learning and ultimately to avoid the curse of multilinguality.

To interpret these incredible results, we introduce the concept of linguistic typology widely used in linguistics as explained in section 2.3. According to the classification of linguistic typology, the degree of word order freedom increases in this order: isolated language, inflectional language, and agglutinative language. The results of our experiments demonstrate that there is a correlation between the freedom of word order and the cross lingual transfer learning effect, which means that including agglutinative language in training pairs results in better performance.

[3] https://huggingface.co/datasets/stsb_multi_mt

It has been commonly considered using languages with structural similarity to improve the cross lingual transfer learning. However, our results show that it is better to consider the freedom of word order, and using agglutinative languages, which have high degree of free word order, not only leads to better performance but also makes our model not fall under the curse of multilinguality.

Based on our work, we contribute to introducing a new perspective on language selection that increases the effectiveness of cross lingual learning.

**Table 1 It shows the experiment results for each learned data. En(E), De(D), Ko(K), and Tr(T) are English, German, Korean, and Turkish, respectively. Performance is reported by convention as spearman correlation × 100.**

| | E-E | D-D | T-T | K-K | E-D | E-T | E-K | D-T | D-K | T-K | Avg |
|---|---|---|---|---|---|---|---|---|---|---|---|
| XLM-RoBERTa | 33.7 | 39.2 | 34.9 | 40.1 | 19.1 | 11.8 | 19.0 | 7.1 | 14.9 | 12.8 | 23.3 |
| Mono En-En | 84.1 | 81.2 | 75.1 | 76.8 | 76.7 | 70.0 | 70.9 | 68.7 | 68.3 | 65.7 | 73.7 |
| Mono De-De | 84.6 | 82.0 | 75.9 | 80.6 | 78.6 | 72.1 | 72.3 | 70.6 | 70.4 | 66.3 | 75.3 |
| Mono Tr-Tr | 85.4 | 82.0 | 77.1 | 81.1 | 78.5 | 73.3 | 73.5 | 71.7 | 71.4 | 68.5 | 76.2 |
| Mono Ko-Ko | 85.3 | 82.1 | 76.2 | 81.4 | 78.8 | 72.9 | 75.0 | 71.4 | 72.3 | 68.4 | **76.4** |
| Cross En-De | 82.8 | 80.5 | 79.5 | 74.0 | 79.6 | 72.8 | 75.3 | 71.8 | 73.6 | 69.6 | 75.9 |
| Cross En-Tr | 83.4 | 79.9 | 79.1 | 74.8 | 79.8 | 76.4 | 76.4 | 74.2 | 73.8 | 71.1 | 76.9 |
| Cross En-Ko | 82.1 | 79.6 | 80.1 | 73.4 | 79.2 | 73.5 | 78.1 | 71.6 | 75.7 | 70.8 | 76.4 |
| Cross De-Tr | 83.6 | 79.9 | 78.5 | 75.1 | 80.0 | 76.1 | 76.2 | 74.5 | 74.3 | 71.7 | 77.0 |
| Cross De-Ko | 83.6 | 80.6 | 80.5 | 74.3 | 79.8 | 73.9 | 78.3 | 72.4 | 76.3 | 71.4 | **77.1** |
| Cross Tr-Ko | 82.8 | 79.0 | 79.0 | 74.7 | 78.9 | 75.3 | 77.5 | 73.2 | 75.1 | 72.7 | 76.8 |
| Mono+Cross En-De | 84.2 | 82.0 | 80.5 | 75.8 | 80.2 | 73.4 | 75.5 | 72.3 | 73.7 | 69.8 | 76.7 |
| Mono+Cross En-Tr | 83.7 | 80.7 | 79.6 | 76.2 | 80.1 | 76.6 | 75.9 | 75.0 | 74.0 | 71.5 | 77.3 |
| Mono+Cross En-Ko | 84.0 | 80.9 | 80.9 | 74.8 | 79.8 | 73.8 | 78.6 | 72.3 | 76.1 | 70.9 | 77.2 |
| Mono+Cross De-Tr | 83.8 | 81.2 | 79.3 | 75.8 | 79.9 | 75.6 | 75.3 | 74.5 | 73.8 | 70.7 | 77.0 |
| Mono+Cross De-Ko | 83.6 | 81.3 | 80.6 | 75.7 | 79.9 | 73.6 | 77.1 | 73.0 | 76.2 | 71.1 | 77.2 |
| Mono+Cross Tr-Ko | 84.3 | 80.9 | 81.0 | 76.5 | 80.3 | 76.4 | 78.0 | 74.4 | 75.9 | 73.3 | **78.1** |
| Whole | 83.0 | 80.3 | 80.3 | 75.1 | 77.7 | 78.1 | 78.9 | 76.0 | 76.0 | 73.0 | **77.8** |

### *Additional Experiments*

Reimers and Gurevych (2019) suggest that the meaning of sentences can be well captured by additionally training with STS dataset on Siamese Network. After we train the base model (NLI_En+Ko) with En, Ko MNLI, Stanford NLI (SNLI, Bowman et al., 2015; Ham et al., 2020) datasets, we create three more models by additionally training the base model with STSb_En, STSb_Ko, and STSb_En+Ko, respectively; this time, we cross-combine STSb_En and STSb_Ko to make STSb_En+Ko.

- (NLI_En+Ko) (Base Model)
- (NLI_En+Ko) + (STSb_En)
- (NLI_En+Ko) + (STSb_Ko)
- (NLI_En+Ko) + (STSb_En+Ko)

In this section, we use STS 2017 dataset (Cer et al., 2017) as test data. As presented in the results of additional experiments, the model trained with Korean pairs outperforms the model trained with the other pairs, which follows the tendency of results explained in the section 3.3. Given that the test dataset does not contain Korean, the cross-language transfer learning effect of the agglutinative language is a truly amazing result.

Presented in the table 3, our model achieves the better performance of XLM-R←SBERT-paraphrase which is the current state of the art model. In addition, our model shows better performance than XLM-R-nli-stsb and XLM-R←SBERT-nli-stsb models trained with only NLI and STS dataset.

**Table 2. It shows the results of STS 2017 for previous studies and our additional experiments. As shown in Table 1, models containing an agglutinative language (Ko) show better performance. Performance is reported by convention as spearman correlation × 100. The other results are referenced from Reimers and Gurebych (2020).**

| | En-En | Es-Es | Ar-Ar | Avg |
|---|---|---|---|---|
| **Ours** | | | | |
| (NLI_En+Ko) (Base Model) | 87.5 | 84.8 | 82.7 | 85.0 |
| (NLI_En+Ko) + (STSb_En) | 88.4 | 86.2 | 81.9 | 85.5 |
| (NLI_En+Ko) + (STSb_Ko) | 88.1 | 87.0 | 82.1 | **85.7** |
| (NLI_En+Ko) + (STSb_En+Ko) | 87.2 | 86.8 | 79.6 | 84.5 |
| **Knowledge Distillation** (Reimers and Gurevych, 2020) | | | | |
| XLMR-nli-stsb | 78.2 | 83.1 | 64.4 | 75.2 |
| XLM-R ← SBERT-nli-stsb | 82.5 | 83.5 | 79.9 | 82.0 |
| XLM-R ← SBERT-paraphrases | 88.8 | 86.3 | 79.6 | **84.9** |
| **Other System** | | | | |
| LASER (Artetxe and Schwenk, 2019) | 77.6 | 79.7 | 68.9 | 75.4 |
| mUSE (Chidambaram et al., 2019) | 86.4 | 86.9 | 76.4 | 83.2 |
| LaBSE (Feng et al., 2020) | 79.4 | 80.8 | 69.1 | 76.4 |

**Table 3 It shows the results of STS2017 for previous studies and our additional experiments. As shown in Table 1, models containing an agglutinative language (Ko) show better performance. Performance is reported by convention as spearman correlation × 100. The other results are referenced from Reimers and Gurebych (2020).**

| | En-Ar | En-De | En-Tr | En-Es | En-Fr | En-It | En-Nl | Avg |
|---|---|---|---|---|---|---|---|---|
| **Ours** | | | | | | | | |
| (NLI_En+Ko) (Base Model) | 73.4 | 84.4 | 71.7 | 82.4 | 80.0 | 79.8 | 84.2 | 79.4 |
| (NLI_En+Ko) + (STSb_En) | 71.9 | 86.0 | 71.6 | 80.5 | 80.2 | 79.2 | 83.4 | 79.0 |
| (NLI_En+Ko) + (STSb_Ko) | 73.3 | 85.9 | 71.4 | 83.7 | 81.5 | 79.7 | 85.2 | **80.1** |
| (NLI_En+Ko) + (STSb_En+Ko) | 74.4 | 85.4 | 72.2 | 82.6 | 81.5 | 80.2 | 83.7 | 80.0 |
| **Knowledge Distillation** | | | | | | | | |
| XLMR-nli-stsb | 44.0 | 59.5 | 42.4 | 54.7 | 63.4 | 59.4 | 66.0 | 55.6 |
| XLM-R ← SBERT-nli-stsb | 77.8 | 78.9 | 74.0 | 79.7 | 78.5 | 78.9 | 77.7 | 77.9 |
| XLM-R ← SBERT-paraphrases | 82.3 | 84.0 | 80.9 | 83.1 | 84.9 | 86.3 | 84.5 | **83.7** |
| **Other System** | | | | | | | | |
| LASER | 66.5 | 64.2 | 72.0 | 75.9 | 69.1 | 70.8 | 68.5 | 69.6 |
| mUSE | 79.3 | 82.1 | 75.5 | 79.6 | 82.6 | 84.5 | 84.1 | 81.1 |
| LaBSE | 74.5 | 73.8 | 72.0 | 65.5 | 77.0 | 76.9 | 75.1 | 73.5 |

***Linguistic Interpretation to Results***

Our research is the first study to reveal that using the agglutinative language has a good effect on cross lingual transfer learning. We assume that relatively high word order freedom of agglutinative languages leads to the performance gains.

According to Transformational Generative Grammar initiated by Noam Chomsky, all humans have the same structures involved in processing speech and language, allowing all languages to share universal attributes. Based on these universal attributes, all languages share semantic representations, which means they all consist of deep structures. At the same time, all the languages have been transformed differently in terms of grammatical, lexical, and phonetic representations, which means they all have different degrees of transformation in surface structures. In light of this aspect, the more transformations languages have, the higher freedom of word order they have. Thus, since they share universal attributes with other languages and learn more transformations by their nature, agglutinative languages lead to better cross lingual transfer learning.

## Conclusion

Multilingual language models have the curse of multilinguality. To mitigate this, it is important to select a language which has a good effect on cross lingual transfer learning.

In general, it is known that the similar linguistic structure is a key factor in performance gains of the cross lingual transfer learning. However, we have demonstrated that agglutinative languages, which allow cross language models to learn phenomenal diversity due to its free word order, are more effective in cross lingual transfer learning.

Finally, we describe some limitations and suggest directions for future work. The first one is that we have not considered all the isolated, inflectional, and agglutinative languages, so to generalize our hypothesis, the future work is required to cross validate more languages. The other limitation is that our hypothesis is attested as a correlation not as a causality; thus, this correlation should be discussed thoroughly by linguists.

## Acknowledgements

“This work is financially supported by Korea Ministry of Land, Infrastructure and Transport (MOLIT) as「Innovative Talent Education Program for Smart City」".